\documentclass{article}
\usepackage{PRIMEarxiv}

\usepackage[T1]{fontenc}    
\usepackage{hyperref}       
\usepackage{url}            
\usepackage{booktabs}       
\usepackage{amsfonts}       
\usepackage{nicefrac}       
\usepackage{microtype}      
\usepackage{amsmath}        
\usepackage{subcaption}
\usepackage{lipsum}
\usepackage{fancyhdr}       
\usepackage{graphicx}  
\usepackage[braket, qm]{qcircuit}
\graphicspath{{media/}}     
\usepackage{tabularx}   
\usepackage{ragged2e}   
\usepackage{threeparttable}

\title{Hybrid Quantum Transformer for Language Generation
\thanks{\textit{\underline{Citation}}: 
    \textbf{Kong. HyQuT.}} 
}

\author{
  Desheng Kong \\
  Nankai University \\
  \texttt{\{kongds@mail.nankai.edu.cn\}} \\
   \And
  Xiangshuo Cui \\
  Nankai University \\
  \texttt{\{2120250604@mail.nankai.edu.cn\}} \\
  \And
  Jiaying Jin \\
  Nankai University \\
  \texttt{\{2120250515@mail.nankai.edu.cn\}} \\
  \AND
  Jing Xu \\
  Nankai University \\
  \texttt{\{xujing@nankai.edu.cn\}} \\
  \And
  Donglin Wang \\
  Nankai University \\
  Beijing Sursen Information Technology Co., Ltd \\
  \texttt{\{wangdonglin@sursen.net\}} \\
}

\begin{document}
\maketitle

\begin{abstract}
Although quantum computing has been increasingly applied to replace classical computation, most existing quantum or hybrid models remain confined to simple tasks, with no successful application to large-scale natural language generation to date. In this work, we present the first hybrid quantum–classical large language model (LLM) for natural language generation, HyQuT, capable of performing coherent and context-aware dialogue. The proposed architecture integrates variational quantum circuits (VQCs) into the Transformer framework at both 8M and 150M parameter scales. Experimental results show that a minimal number of qubits (10 qubits with 80 quantum gates) can replace about 10\% of the classical parameters in the 150M-parameter model, while achieving comparable convergence stability and generation quality. This study provides an early demonstration of the feasibility of integrating quantum computing to large-scale generative language models.
\end{abstract}


\section{Introduction}
Recent advances in Transformer-based large language models (LLMs), beginning with the seminal Transformer architecture\cite{NIPS2017_3f5ee243}, have driven remarkable progress in natural language generation. The paradigm of scaling these models, exemplified by landmark architectures like the GPT series\cite{radford2019language,brown2020language} and now widely adopted in open-source models such as LLaMA\cite{touvron2023llama}, has unlocked unprecedented capabilities. Yet, this continual scaling has led to substantial and often prohibitive computational and energy demands\cite{patterson2021carbon}. As model parameters grow into the hundreds of millions or billions, training and inference increasingly strain classical hardware resources, highlighting the urgent need for new computing paradigms to alleviate these bottlenecks. Quantum computing, with its potential for exponential parallelism and compact information representation, has emerged as a promising alternative for enhancing neural architectures\cite{biamonte2017quantum}.

Despite rapid progress in quantum machine learning, most existing quantum or hybrid models remain limited to small-scale tasks such as classification, regression, or time-series forecasting. Their architectures typically involve shallow quantum circuits or isolated quantum layers without integration into large-scale generative models. To date, no study has successfully demonstrated a quantum–classical hybrid model capable of coherent and context-aware language generation, leaving an unexplored gap between quantum computation and large-scale natural language processing (NLP).

In this work, we present the first hybrid quantum–classical LLM designed for generative NLP tasks, HyQuT. Our approach explores the feasibility of embedding variational quantum circuits (VQCs) into different layers of Transformer architectures to replace part of the classical linear transformations, thereby reducing parameter redundancy while preserving generative capability. This hybrid integration enables the model to leverage quantum representations for feature projection and contextual reasoning within the Transformer framework.

We evaluate HyQuT at both 8M and 150M parameter scales through pre-training and fine-tuning experiments on high-quality Chinese corpora. Results show that approximately 10\% of the classical parameters can be replaced by a minimal number of quantum resources—only 10 qubits with 80 quantum gates—while achieving comparable convergence stability and text generation quality. These findings demonstrate that even under near-term hardware constraints, quantum circuits can provide tangible benefits to large-scale generative models.

In summary, this study contributes (1) the first demonstration of a hybrid quantum–classical large model capable of natural language generation, and (2) empirical evidence that quantum circuits can effectively substitute part of the classical computation in Transformers with reduced parameters and comparable performance, providing a feasible direction for quantum-enhanced NLP models.

\section{Related Works}

\subsection{Large Language Models and the Scaling Paradigm}
The foundation of modern NLP was revolutionized by the Transformer architecture, which introduced the self-attention mechanism as a powerful tool for capturing long-range dependencies in text \cite{NIPS2017_3f5ee243}. This design departed from recurrent structures, enabling massive parallelization and paving the way for models of unprecedented scale. Early applications of this architecture, such as BERT \cite{devlin2019bert}, demonstrated the power of large-scale pre-training on general language understanding tasks, typically followed by fine-tuning for specific applications.

A subsequent paradigm shift moved towards generative pre-training, where models are trained to predict the next token in a sequence. This approach, championed by the GPT family of models \cite{radford2019language,brown2020language}, revealed that scaling up model size, dataset size, and compute leads to predictable improvements in performance, a phenomenon formalized as "scaling laws"\cite{kaplan2020scaling}. These (LLMs) exhibit emergent abilities, allowing them to perform a wide range of tasks with zero or few-shot prompting, without task-specific training. The development of instruction tuning \cite{ouyang2022training} and reinforcement learning from human feedback (RLHF) \cite{christiano2017deep} further refined these models' ability to align with user intent, making them powerful and steerable tools.

The contemporary LLM landscape is characterized by both enormous proprietary models and a burgeoning ecosystem of powerful open-source alternatives like LLaMA and its derivatives \cite{touvron2023llama,touvron2023llama2}. While their capabilities are vast, the immense computational cost remains a central challenge. In response, the classical machine learning community has developed various parameter-efficient fine-tuning (PEFT) techniques, such as LoRA \cite{hu2022lora}, to reduce the cost of adapting these models. While such methods address the adaptation phase, the fundamental cost of pre-training and inference remains. This persistent challenge motivates exploring alternative computational paradigms, such as quantum computing, to fundamentally reimagine the building blocks of these powerful architectures.

\subsection{Quantum Natural Language Processing (QNLP)}
Quantum Natural Language Processing (QNLP) has emerged as a novel paradigm aiming to leverage the principles of quantum mechanics to model the complexities of human language. Early foundational work in this area, notably the DisCoCat framework, proposed using the compositional structure of quantum tensor networks to represent the grammatical and semantic composition of sentences \cite{coecke2010mathematical}. This approach holds theoretical appeal, as the high-dimensional Hilbert spaces and phenomena like superposition and entanglement could offer more natural and resource-efficient representations for complex linguistic features, such as semantic ambiguity and logical entailment.

However, these early QNLP models, while theoretically elegant, have largely been confined to syntactic parsing or small-scale semantic classification tasks. Their architectures are typically rigid and task-specific, lacking the scalability and, crucially, the generative capabilities that characterize modern, large-scale language models. To bridge this gap and unlock the potential of QNLP for more complex applications, the research community has increasingly turned its attention toward developing quantum analogues of the highly successful Transformer architecture.

\subsection{Quantum Self-Attention and Quantum Transformers}
The cornerstone of the Transformer model is its self-attention mechanism. Consequently, the primary challenge in creating a quantum Transformer lies in designing an efficient and physically realizable Quantum Self-Attention (QSA) module. This has spurred a variety of innovative approaches, which, despite their architectural diversity, reveal a striking convergence in their ultimate application goals.

The initial and most direct line of inquiry has focused on creating a quantum analogue of the classical self-attention process. The pioneering QSANN model first introduced a framework where Query, Key, and Value vectors are encoded into parameterized quantum circuits (PQCs), and their inner-product similarity is estimated via quantum measurements \cite{li2024quantum}. The viability of this core concept was successfully demonstrated on several text classification benchmarks. Following this, a significant body of work has sought to refine and extend this paradigm. These advancements include enhancing measurement schemes to reduce parameter counts \cite{zhang2024light}, incorporating more generalized measurement techniques like POVMs \cite{wei2023povm}, and adapting the architecture for different data modalities. Notably, this QSA mechanism has been extended from its NLP origins to the domains of computer vision \cite{li2024quantum,zhang2025hqvit}, high-energy physics \cite{unlu2024hybrid,comajoan2024quantum}, and time-series forecasting \cite{chakraborty2025integrating,dutta2024qadqn}. These efforts, while expanding the applicability of QSA, have consistently targeted problems of a discriminative nature, where the objective is to classify, predict, or regress a specific output.

A parallel research thrust has explored more sophisticated methods for computing attentional similarity, moving beyond simple inner products. For instance, some models have proposed novel metrics such as quantum logical similarity \cite{shi2024qsan} or complex-valued similarity that leverages the phase information of qubits \cite{chen2025quantum} to capture non-linear relationships between inputs. Other novel designs, such as holistic attention models that bypass explicit score computation \cite{kerenidis2024quantum} have also been proposed. Yet, their evaluation remains within the confines of classification or theoretical analysis.

This comprehensive review reveals an unmistakable pattern: the overwhelming body of research on quantum Transformers is dedicated to discriminative tasks. The fundamental goal of these models is to compress high-dimensional input data into a distilled representation suitable for making a classification or prediction. To the best of our knowledge, the crucial domain of generative tasks, particularly in NLP, remains entirely unexplored. This represents a significant omission, as generative capabilities are a hallmark of modern artificial intelligence and a critical step towards achieving more sophisticated and human-like language understanding and creation.

To address this critical gap, this paper introduces the first quantum Transformer architecture designed and implemented specifically for generative NLP. Our work aims to expand the horizon of QNLP from discriminative modeling to the more challenging and impactful realm of language generation.

\section{Method}

\subsection{Quantum-Classical Hybrid Computational Paradigm Architecture Design}
The core of HyQuT is an innovative quantum-classical hybrid module, engineered to function as a direct replacement for standard linear projection layers and to transcend the expressive limitations of classical linear transformations. By deeply integrating a variational quantum circuit into the Transformer architecture, we leverage the intrinsic superposition and entanglement properties of quantum states. This allows the module to operate within and explore a feature space of exponential capacity—a Hilbert space scaling as $\mathcal{O}(2^{n_q})$, where ${n_q}$is the number of qubits. This stands in stark contrast to a traditional parameter matrix, such as the Query projection  $\mathbf{W}^Q \in \mathbb{R}^{d_{model} \times d_k}$, whose expressive power is constrained by a polynomial number of real-valued parameters. Through this integration, we enable end-to-end differentiable optimization of complex, non-linear feature transformations, creating a theoretical path for quantum-enhanced representation learning in large-scale models.

However, harnessing this theoretical power within a practical hybrid system presents a formidable challenge: the seamless integration of a high-dimensional classical feature space (e.g., a model's hidden dimension of 512) with the constrained, low-dimensional nature of near-term quantum processors (e.g., 10-20 qubits). A naive, direct encoding is infeasible with current hardware and algorithms. To address this challenge, our module introduces a practical and scalable three-stage architecture: (1) a classical dimensionality reduction encoder, (2) a core parameterized quantum circuit, and (3) a classical dimensionality augmentation decoder. This structure effectively bridges the dimensional gap, making the integration of quantum circuits into large-scale models tractable while allowing the entire module to be trained end-to-end within a standard deep learning framework.

\begin{figure}
  \centering
  \includegraphics[width=1\textwidth]{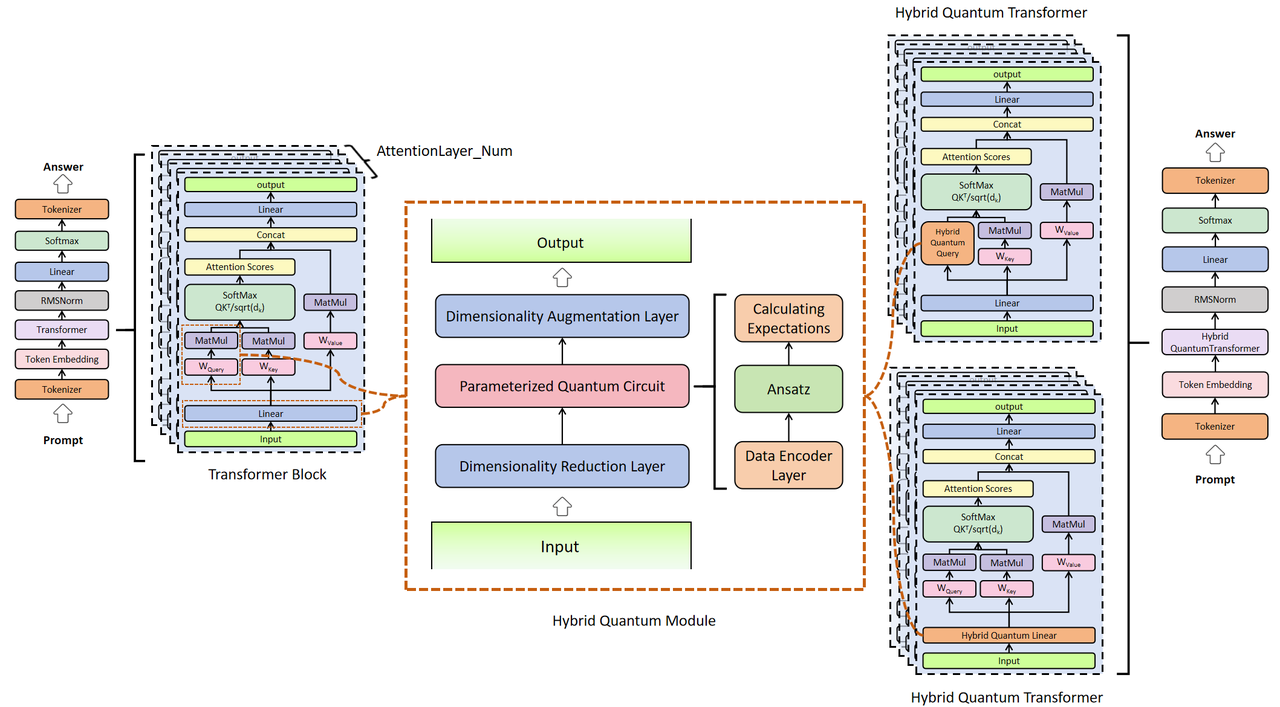}
  \caption{The architecture of the Overall Hybrid Architecture. The framework consists of a general-purpose projection module (middle) and its specific integration into the Transformer architecture (right).}
  \label{fig:overall-architecture}
\end{figure}

In Figure \ref{fig:overall-architecture}, the Hybrid Quantum Module, framed by the orange dotted line, is a general-purpose projection module. It takes a high-dimensional classical vector Input from the Transformer. A Dimensionality Reduction Layer first compresses the input into a low-dimensional representation suitable for the quantum backend. This vector is then encoded into a parameterized quantum circuit, which performs a variational transformation. Finally, the measurement result from the quantum circuit is fed into a Dimensionality Augmentation Layer to project it back to the original hidden dimension, producing the final Output vector. The entire module acts as a drop-in replacement for a classical linear layer. The top-right diagram illustrates the architecture of HyQuT-150M, which replaces the Query projection matrix ($W_q$) in the self-attention block, while the bottom-right diagram shows the architecture for HyQuT-8M, where the gate projection matrix in the FFN block is replaced.

\subsection{Quantum-Adaptive Dimensionality Reduction for High-Dimensional Semantic Space}
Given the input semantic representation tensor $\mathbf{X}^{(\ell)} \in \mathbb{R}^{B \times L \times d_{model}}$ at the$\ell$-th Transformer layer, where $B \in \mathbb{Z}^+$ represents the mini-batch training sample size, $L \in \mathbb{Z}^+$ denotes the input sequence token length, and $d_{model} \in \mathbb{Z}^+$ indicates the Transformer hidden layer feature dimension. To achieve effective mapping from classical high-dimensional feature spaces to quantum Hilbert spaces, we introduce an Adaptive Feature Compression Module that projects semantic vectors into quantum circuit-processable subspaces through learnable affine transformations:
\begin{equation}
 \mathbf{X}_{compressed}^{(\ell)} = \sigma_{compress}\left(\mathbf{X}^{(\ell)} \mathbf{W}_{down}^{(\ell)} + \mathbf{b}_{down}^{(\ell)}\right) \tag{1}
\end{equation}
where the dimensionality reduction projection matrix $\mathbf{W}_{down}^{(\ell)} \in \mathbb{R}^{d_{model} \times 2n_q}$ and bias vector $\mathbf{b}_{down}^{(\ell)} \in \mathbb{R}^{2n_q}$are trainable parameters for this layer, and $\sigma_{compress}(\cdot)$ is the Tanh activation function ensuring feature value boundedness. The compressed tensor $\mathbf{X}_{compressed}^{(\ell)} \in \mathbb{R}^{B \times L \times 2n_q}$preserves the core information entropy of the original semantic representation while reducing dimensionality to a $2n_q$-dimensional submanifold matching quantum encoding requirements.

The $2n_q$-dimensional design is based on the dual degree-of-freedom requirement for quantum state encoding: each qubit requires two independent rotation angle parameters$(\theta, \phi)$ to fully determine its quantum state on the Bloch sphere. This design ensures injective mapping from classical information to quantum states, preventing information loss.

\subsection{Tensor Reconstruction Operator for Quantum Circuit Batch Processing}
To enable efficient batched processing of quantum circuits, we define a tensor reconstruction operator $\mathcal{T}_{batch}: \mathbb{R}^{B \times L \times 2n_q} \rightarrow \mathbb{R}^{(B \cdot L) \times 2n_q}$ that merges the batch and sequence dimensions into a unified quantum circuit parallel processing dimension:
\begin{equation}
\mathbf{X}_{flat}^{(\ell)} = \mathcal{T}_{batch}(\mathbf{X}_{compressed}^{(\ell)}) = \text{Flatten}_{[0,1]}(\mathbf{X}_{compressed}^{(\ell)}) \tag{2}
\end{equation}
where $\text{Flatten}_{[0,1]}(\cdot)$denotes the tensor reorganization operation flattening along the first two dimensions. The transformed $\mathbf{X}_{flat}^{(\ell)} \in \mathbb{R}^{(B \cdot L) \times 2n_q}$can be viewed as $N_{batch} = B \cdot L$ independent quantum circuit input instances, each carrying $2n_q$classical feature values to be encoded into quantum states. This tensor reconstruction strategy fully exploits the parallelism of quantum circuit execution, enabling a single quantum computation to simultaneously process all time steps across the entire batch and sequence, significantly enhancing computational efficiency.

\subsection{Parameterized Quantum State Encoding Protocol}
\begin{figure*}[htpb]
    \centering
    \begin{subfigure}[b]{0.9\textwidth}
        \centering
        \Qcircuit @C=1.5em @R=1.2em {
        \lstick{\ket{0}} & \gate{H} & \gate{RY}& \gate{RZ} & \ctrl{1} & \qw & \ctrl{1}  & \qw        & \ctrl{1} & \qw       & \ctrl{1}   & \qw       & \ctrl{1} & \qw     & \gate{RY}&\meter\\
        \lstick{\ket{0}} & \gate{H} & \gate{RY} & \gate{RZ} & \targ    & \qw & \gate{RY}  & \ctrl{1}  & \targ    & \ctrl{1}  & \gate{RY}  & \ctrl{1}  & \targ    & \ctrl{1}& \gate{RY}&\meter\\
        \lstick{\ket{0}} & \gate{H} & \gate{RY} & \gate{RZ} & \ctrl{1} & \qw & \ctrl{1}  & \gate{RZ}  & \ctrl{1} & \targ     & \ctrl{1}   & \gate{RZ}  & \ctrl{1} & \targ  & \gate{RY}&\meter\\
        \lstick{\ket{0}} & \gate{H} & \gate{RY} & \gate{RZ} & \targ    & \qw & \gate{RY}  & \ctrl{1}  & \targ    & \ctrl{1}  & \gate{RY}  & \ctrl{1}  & \targ    & \ctrl{1}& \gate{RY}&\meter\\
        \lstick{\ket{0}} & \gate{H} & \gate{RY} & \gate{RZ} & \ctrl{1} & \qw & \ctrl{1}  & \gate{RZ}  & \ctrl{1} & \targ     & \ctrl{1}   & \gate{RZ}  & \ctrl{1} & \targ  & \gate{RY}&\meter\\
        \lstick{\ket{0}} & \gate{H} & \gate{RY} & \gate{RZ} & \targ    & \qw & \gate{RY}  & \ctrl{1}  & \targ    & \ctrl{1}  & \gate{RY}  & \ctrl{1}  & \targ    & \ctrl{1}& \gate{RY}&\meter\\
        \lstick{\ket{0}} & \gate{H} & \gate{RY} & \gate{RZ} & \ctrl{1} & \qw & \ctrl{1}  & \gate{RZ}  & \ctrl{1} & \targ     & \ctrl{1}   & \gate{RZ}  & \ctrl{1} & \targ  & \gate{RY}&\meter\\
        \lstick{\ket{0}} & \gate{H} & \gate{RY} & \gate{RZ} & \targ    & \qw & \gate{RY}  & \ctrl{1}  & \targ    & \ctrl{1}  & \gate{RY}  & \ctrl{1}  & \targ    & \ctrl{1}& \gate{RY}&\meter\\
        \lstick{\ket{0}} & \gate{H} & \gate{RY} & \gate{RZ} & \ctrl{1} & \qw & \ctrl{1}  & \gate{RZ}  & \ctrl{1} & \targ     & \ctrl{1}   & \gate{RZ}  & \ctrl{1} & \targ  & \gate{RY}&\meter\\
        \lstick{\ket{0}} & \gate{H} & \gate{RY} & \gate{RZ} & \targ    & \qw & \gate{RY} & \qw  & \targ   & \qw  & \gate{RY}  & \qw  & \targ   & \qw & \gate{RY}&\meter\gategroup{1}{6}{10}{15}{1.1em}{--}\\
        & & &  & && &  &  &   &   &  &  &   & \dstick{ \times L}&
        }
        \caption{} 
        \label{fig:suba}
    \end{subfigure}
    
    \begin{subfigure}[b]{0.9\textwidth}
        \centering
        \Qcircuit @C=2.2em @R=1.2em {
        \lstick{\ket{0}} & \gate{H} & \gate{RY}& \gate{RZ} & \ctrl{1} & \qw & \gate{R_Z} & \gate{R_Y} & \gate{R_Z}        & \ctrl{1} & \qw    & \gate{RY}&\meter\\
        \lstick{\ket{0}} & \gate{H} & \gate{RY} & \gate{RZ} & \targ    & \qw & \gate{R_Z} & \gate{R_Y} & \gate{R_Z}  & \targ    & \ctrl{1}  & \gate{RY}&\meter\\
        \lstick{\ket{0}} & \gate{H} & \gate{RY} & \gate{RZ} & \ctrl{1} & \qw & \gate{R_Z} & \gate{R_Y} & \gate{R_Z}  & \ctrl{1} & \targ       & \gate{RY}&\meter\\
        \lstick{\ket{0}} & \gate{H} & \gate{RY} & \gate{RZ} & \targ    & \qw & \gate{R_Z} & \gate{R_Y} & \gate{R_Z}  & \targ    & \ctrl{1}  & \gate{RY}&\meter\\
        \lstick{\ket{0}} & \gate{H} & \gate{RY} & \gate{RZ} & \ctrl{1} & \qw & \gate{R_Z} & \gate{R_Y} & \gate{R_Z}  & \ctrl{1} & \targ     & \gate{RY}&\meter\\
        \lstick{\ket{0}} & \gate{H} & \gate{RY} & \gate{RZ} & \targ    & \qw & \gate{R_Z} & \gate{R_Y} & \gate{R_Z} & \targ    & \ctrl{1}  & \gate{RY}&\meter\\
        \lstick{\ket{0}} & \gate{H} & \gate{RY} & \gate{RZ} & \ctrl{1} & \qw & \gate{R_Z} & \gate{R_Y} & \gate{R_Z} & \ctrl{1} & \targ     & \gate{RY}&\meter\\
        \lstick{\ket{0}} & \gate{H} & \gate{RY} & \gate{RZ} & \targ    & \qw & \gate{R_Z} & \gate{R_Y} & \gate{R_Z} & \targ    & \ctrl{1} & \gate{RY}&\meter\\
        \lstick{\ket{0}} & \gate{H} & \gate{RY} & \gate{RZ} & \ctrl{1} & \qw & \gate{R_Z} & \gate{R_Y} & \gate{R_Z}  & \ctrl{1} & \targ     & \gate{RY}&\meter\\
        \lstick{\ket{0}} & \gate{H} & \gate{RY} & \gate{RZ} & \targ    & \qw & \gate{R_Z} & \gate{R_Y} & \gate{R_Z}  & \targ   & \qw  & \gate{RY}&\meter\gategroup{1}{6}{10}{11}{2.5em}{--}\\
        & & &  & & & &  &  &    & \dstick{ \times L}&
        }
        \caption{} 
        \label{fig:subb}
    \end{subfigure}
    \caption{A schematic diagram of quantum circuits, where (a) is the quantum circuit of the HyQuT-8M model and (b) is the quantum circuit of the HyQuT-150M. The main difference between the two is the entanglement strength of the variable quantum parameters and the type of rotating gate.}
    \label{fig:quantum-circuits}
\end{figure*}
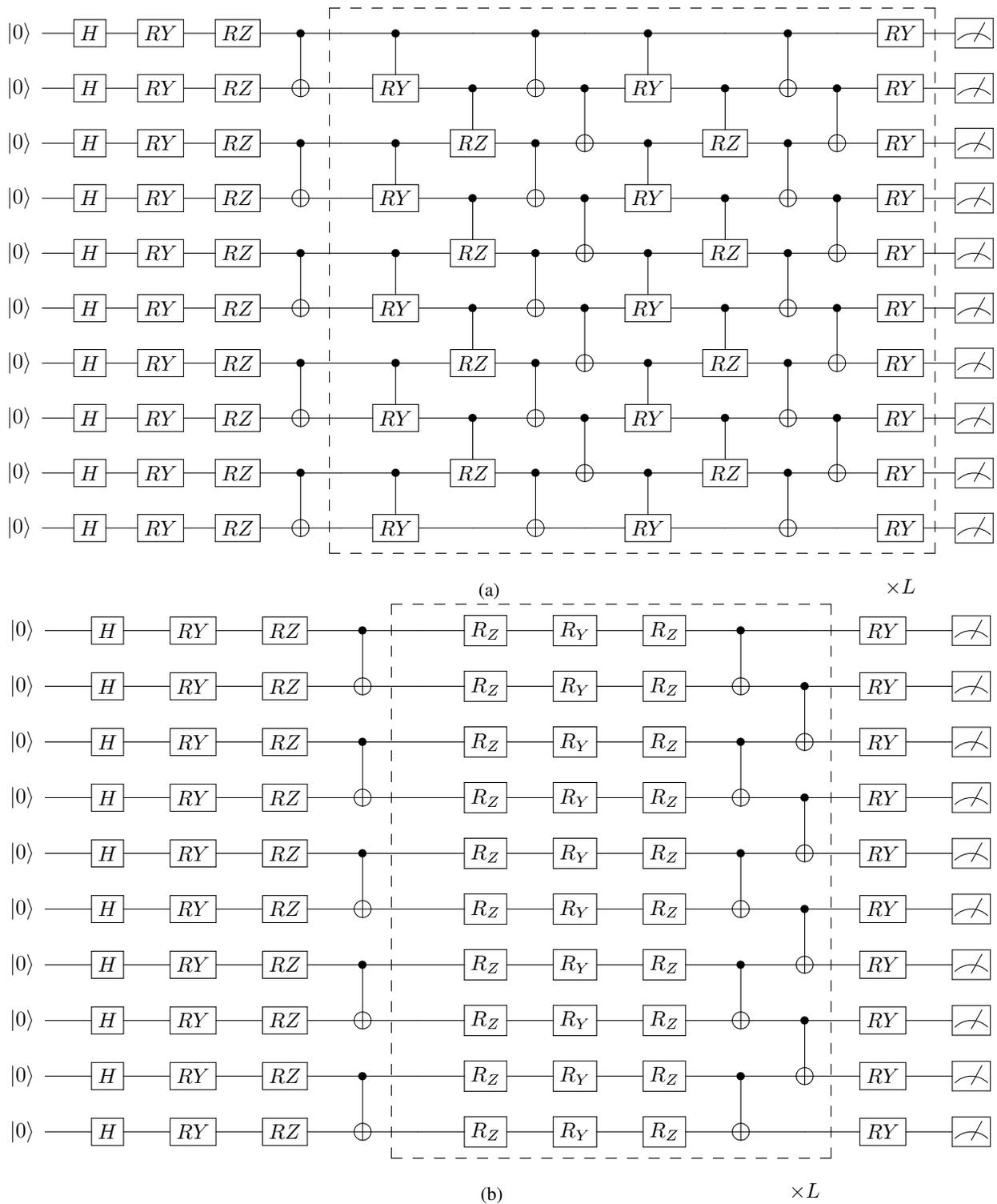

For the $i$-th input instance $\mathbf{x}_i = [x_{i,1}, x_{i,2}, \ldots, x_{i,2n_q}]^T \in \mathbb{R}^{2n_q}$where $i \in {1, 2, \ldots, N_{batch}}$, we construct a quantum register containing $n_q$qubits and employ a three-stage parameterized encoding protocol to map classical features to quantum states. Two quantum circuits used in the model are shown in Figure \ref{fig:quantum-circuits}, with Figure \ref{fig:subb} as an example.

\subsubsection{Stage I: Uniform Initialization of Quantum State Space}
The quantum register is initialized to the ground state $|0\rangle^{\otimes n_q}$, followed by application of Hadamard gates $H$ to all qubits to construct a uniform superposition of maximally mixed states:
\begin{equation}
    |\psi_0\rangle = \mathcal{H}^{\otimes n_q}|0\rangle^{\otimes n_q} = \bigotimes_{j=1}^{n_q} \frac{1}{\sqrt{2}}(|0\rangle_j + |1\rangle_j) = \frac{1}{\sqrt{2^{n_q}}} \sum_{k=0}^{2^{n_q}-1} |k\rangle \tag{3} 
\end{equation}
where $\mathcal{H}$ represents the Hadamard gate operator, defined as:
\begin{equation}
    H = \frac{1}{\sqrt{2}}\begin{pmatrix} 1 & 1 \\ 1 & -1 \end{pmatrix} \tag{4} 
\end{equation}
This initialization creates an equal-probability quantum superposition of $2^{n_q}$ computational basis states, providing maximal state space exploration potential for subsequent feature encoding and embodying the essential advantage of quantum parallelism.

\subsubsection{Stage II: Angular Embedding for Feature Encoding}
For qubit $j \in {1, 2, \ldots, n_q}$, parameterized rotation gates $R_Y(\theta_j)$ and $R_Z(\phi_j)$ are sequentially applied around the Y and Z axes, implementing dual-channel encoding of classical features into quantum state amplitudes and phases:
\begin{equation}
    R_Y(\theta) = \exp\left(-i\frac{\theta}{2}\hat{\sigma}_y\right) = \begin{pmatrix} \cos(\theta/2) & -\sin(\theta/2) \\ \sin(\theta/2) & \cos(\theta/2) \end{pmatrix} \tag{5} 
\end{equation}
\begin{equation}
    R_Z(\phi) = \exp\left(-i\frac{\phi}{2}\hat{\sigma}_z\right) = \begin{pmatrix} e^{-i\phi/2} & 0 \\ 0 & e^{i\phi/2} \end{pmatrix} \tag{6} 
\end{equation}
where $\hat{\sigma}_y, \hat{\sigma}_z$ are Pauli operators. Rotation angles are extracted from input features through nonlinear mapping functions:
\begin{equation}
    \theta_j = \pi \cdot \sigma(x_{i,j}), \quad \phi_j = \pi \cdot \sigma(x_{i,n_q+j}) \tag{7} 
\end{equation}
Here $\sigma(z) = (1 + e^{-z})^{-1}$ is the Sigmoid function with range $[0,1]$, scaled by $\pi$ to ensure rotation angles lie within the $[0, \pi]$ interval, covering the southern hemisphere of the Bloch sphere and providing sufficient state space representational capacity.

\subsubsection{Stage III: Quantum Superposition of Encoded States}
After completing angular encoding, the quantum state corresponding to the $i$-th input instance is the tensor product of individual qubit encoded states:
\begin{equation}
    \ket{\psi_{\text{enc}}^{(i)}} = \bigotimes_{j=1}^{n_q} R_Z(\phi_j) R_Y(\theta_j) H \ket{0}_j \tag{8}
\end{equation}
This quantum state forms a high-dimensional quantum embedding of the input features in the $2^{n_q}$-dimensional Hilbert space $\mathcal{H}^{\otimes n_q}$, with its state vector containing $2^{n_q}$ complex amplitude components, far exceeding the information capacity of classical $2n_q$-dimensional inputs and demonstrating the exponential expressive advantage of quantum encoding.

\subsection{Hierarchical Variational Quantum Feature Transformation}
The encoded quantum state undergoes deep feature extraction and nonlinear transformation through $N_L \in \{2, 3\}$ layers of parameterized Ansatz circuits. Each Ansatz layer constitutes a unitary transformation operator, with design following the Hardware-Efficient Ansatz paradigm, balancing expressive power and near-term quantum device realizability.

\subsubsection{Unitary Decomposition of Single-Layer Ansatz}
The $\ell$-th layer Ansatz operator $U_{\text{ansatz}}^{(\ell)}: \mathcal{H}^{\otimes n_q} \rightarrow \mathcal{H}^{\otimes n_q}$ can be decomposed into the composition of a parameterized rotation layer $U_{\text{rot}}^{(\ell)}$ and an entangling gate layer $U_{\text{ent}}^{(\ell)}$:
\begin{equation}
    U_{\text{ansatz}}^{(\ell)} = U_{\text{ent}}^{(\ell)} \circ U_{\text{rot}}^{(\ell)} \tag{9}
\end{equation}
The parameterized rotation layer consists of independent three-rotation gate sequences on each qubit:
\begin{equation}
    U_{\text{rot}}^{(\ell)} = \bigotimes_{j=1}^{n_q} R_Z(\alpha_j^{(\ell)}) R_Y(\beta_j^{(\ell)}) R_Z(\gamma_j^{(\ell)}) \tag{10}
\end{equation}
where the variational parameter set $\Theta^{(\ell)} = \{\alpha_j^{(\ell)}, \beta_j^{(\ell)}, \gamma_j^{(\ell)}\}_{j=1}^{n_q}$ is optimized through end-to-end training. This $ZYZ$ decomposition enables universal approximation of arbitrary single-qubit unitary gates on the Bloch sphere, ensuring expressive completeness at the single-qubit level.

\subsubsection{Quantum Entanglement Induction Mechanism}
The entangling gate layer $U_{\text{ent}}^{(\ell)}$ establishes quantum correlations between adjacent qubits through controlled-NOT (CNOT) gates:
\begin{equation}
    U_{\text{ent}}^{(\ell)} = \prod_{j=1}^{n_q} \text{CNOT}_{j \rightarrow [(j \bmod n_q) + 1]} \tag{11}
\end{equation}
where $\text{CNOT}_{c \rightarrow t}$ denotes a controlled-NOT gate with control qubit $c$ acting on target qubit $t$, with matrix representation:
\begin{equation}
    \text{CNOT} = \ket{0}\bra{0} \otimes \mathbb{I} + \ket{1}\bra{1} \otimes \hat{X} = \begin{pmatrix} 1 & 0 & 0 & 0 \\ 0 & 1 & 0 & 0 \\ 0 & 0 & 0 & 1 \\ 0 & 0 & 1 & 0 \end{pmatrix} \tag{12}
\end{equation}
The circular connectivity topology $(1 \rightarrow 2 \rightarrow \cdots \rightarrow n_q \rightarrow 1)$ ensures global information propagation among all qubits, generating multipartite entangled states. Quantum entanglement causes the system's joint state to be non-decomposable into simple tensor products of subsystem states, producing nonlocal quantum correlations that transcend classical correlations—a key source of quantum neural network expressive power.

\subsubsection{Deep Quantum Evolution}
After $N_L$ cascaded transformations, the final quantum state is:
\begin{equation}
    \ket{\psi_{\text{final}}^{(i)}} = \mathcal{U}_{\text{VQC}}(\Theta) \ket{\psi_{\text{enc}}^{(i)}} = \left(\prod_{\ell=1}^{N_L} U_{\text{ansatz}}^{(\ell)}\right) \ket{\psi_{\text{enc}}^{(i)}} \tag{13}
\end{equation}
where $\mathcal{U}_{\text{VQC}}(\Theta)$ represents the complete VQC unitary operator, and $\Theta = \bigcup_{\ell=1}^{N_L} \Theta^{(\ell)}$ encompasses all variational parameters across layers. According to the Solovay-Kitaev theorem, as layer number $N_L$ increases and parameter count expands, this hierarchical unitary circuit can approximate arbitrary unitary transformations in the $SU(2^{n_q})$ group with arbitrary precision, endowing VQCs with powerful function approximation capabilities.

The quantum circuit depth $N_L$ selection balances expressive power and training complexity: $N_L=2$ layers can establish fundamental quantum entanglement structures, while $N_L=3$ layers further enhance nonlinear modeling capabilities through additional parameterized degrees of freedom, while controlling gradient computation circuit execution costs.

\subsection{Quantum Observable Measurement and Expectation Value Extraction}
To extract classical information from evolved quantum states for subsequent neural network computation, we perform projective measurements based on Pauli-Z operators on each qubit and compute measurement operator quantum expectation values as output features.

\subsubsection{Construction of Local Observables}
For the $j$-th qubit, define the local measurement operator as:
\begin{equation}
    \hat{O}_j = \mathbb{I}^{\otimes (j-1)} \otimes \hat{\sigma}_z^{(j)} \otimes \mathbb{I}^{\otimes (n_q-j)} \tag{14}
\end{equation}
where the Pauli-Z operator $\hat{\sigma}_z = \ket{0}\bra{0} - \ket{1}\bra{1}$ has eigenvalues $\{+1, -1\}$, corresponding to the qubit being in states $\ket{0}$ and $\ket{1}$ respectively. This diagonalized operator choice enables measurement expectation values to be computed via probability weighting:
\begin{equation}
    m_j^{(i)} = \bra{\psi_{\text{final}}^{(i)}} \hat{O}_j \ket{\psi_{\text{final}}^{(i)}} = P_0^{(j)} - P_1^{(j)} \tag{15}
\end{equation}
where $P_0^{(j)}, P_1^{(j)}$ are the probabilities of measuring the $j$-th qubit in states $\ket{0}, \ket{1}$ respectively. The expectation value $m_j^{(i)} \in [-1, +1]$ encodes statistical information about that qubit's state: $m_j \approx +1$ indicates primarily in state $\ket{0}$, $m_j \approx -1$ indicates primarily in state $\ket{1}$, and $m_j \approx 0$ indicates a maximally mixed state.

\subsubsection{Batched Measurement Protocol}
Performing parallel measurements on all $n_q$ qubits constitutes the output vector for the $i$-th circuit instance:
\begin{equation}
    \mathbf{m}^{(i)} = 
    \begin{bmatrix} 
        m_1^{(i)} \\ m_2^{(i)} \\ \vdots \\ m_{n_q}^{(i)} 
    \end{bmatrix} = 
    \begin{bmatrix} 
        \langle \hat{O}_1 \rangle_{\ket{\psi_{\text{final}}^{(i)}}} \\ 
        \langle \hat{O}_2 \rangle_{\ket{\psi_{\text{final}}^{(i)}}} \\ 
        \vdots \\ 
        \langle \hat{O}_{n_q} \rangle_{\ket{\psi_{\text{final}}^{(i)}}} 
    \end{bmatrix} 
    \in \mathbb{R}^{n_q} \tag{16}
\end{equation}
Executing the above measurement process on all $N_{\text{batch}} = B \cdot L$ quantum circuit instances yields the batched output matrix:
\begin{equation}
    \mathbf{M}_{\text{flat}}^{(\ell)} = 
    \begin{bmatrix} 
        (\mathbf{m}^{(1)})^T \\ 
        (\mathbf{m}^{(2)})^T \\ 
        \vdots \\ 
        (\mathbf{m}^{(N_{\text{batch}})})^T 
    \end{bmatrix} 
    \in \mathbb{R}^{(B \cdot L) \times n_q} \tag{17}
\end{equation}
Each row of this matrix corresponds to quantum feature representations for a timestep-sample pair, containing $n_q$-dimensional quantum measurement statistics that encapsulate high-order feature information processed through quantum entanglement and interference.

\subsection{Inverse Tensor Reconstruction and Feature Space Upscaling}
To re-embed quantum measurement results into the Transformer's sequential structure, we define an inverse tensor reconstruction operator $\mathcal{T}_{\text{batch}}^{-1}: \mathbb{R}^{(B \cdot L) \times n_q} \rightarrow \mathbb{R}^{B \times L \times n_q}$ that restores the separated batch and sequence dimensions:
\begin{equation}
    \mathbf{M}^{(\ell)} = \mathcal{T}_{\text{batch}}^{-1}(\mathbf{M}_{\text{flat}}^{(\ell)}) = \text{Unflatten}_{[0 \rightarrow (0,1)]}(\mathbf{M}_{\text{flat}}^{(\ell)}, (B, L)) \tag{18}
\end{equation}
where $\text{Unflatten}(\cdot)$ denotes the tensor inverse operation reorganizing flattened dimensions into specified shapes.

Subsequently, an Adaptive Feature Expansion Module projects quantum outputs back to the original hidden layer dimension through learnable linear transformations and nonlinear activations:
\begin{equation}
    \mathbf{Q}^{(\ell)} = \text{GELU}\left(\mathbf{M}^{(\ell)} \mathbf{W}_{\text{up}}^{(\ell)} + \mathbf{b}_{\text{up}}^{(\ell)}\right) \tag{19}
\end{equation}
where the upscaling projection matrix $\mathbf{W}_{\text{up}}^{(\ell)} \in \mathbb{R}^{n_q \times d_{\text{model}}}$ and bias vector $\mathbf{b}_{\text{up}}^{(\ell)} \in \mathbb{R}^{d_{\text{model}}}$ are trainable parameters for this layer. The GELU (Gaussian Error Linear Unit) activation function is defined as:
\begin{equation}
    \text{GELU}(x) = x \cdot \Phi(x) = x \cdot \frac{1}{2}\left[1 + \text{erf}\left(\frac{x}{\sqrt{2}}\right)\right] \tag{20}
\end{equation}
where $\Phi(\cdot)$ is the cumulative distribution function of the standard normal distribution, and $\text{erf}(\cdot)$ is the error function. GELU provides smooth nonlinear activation, facilitating gradient flow and feature learning.
The final Query representation $\mathbf{Q}^{(\ell)} \in \mathbb{R}^{B \times L \times d_{\text{model}}}$ completely replaces the traditional self-attention mechanism's linear projection $\mathbf{Q} = \mathbf{X}\mathbf{W}^Q$, while injecting quantum-processed high-order nonlinear features.

\subsection{Quantum-Enhanced Multi-Head Self-Attention Mechanism}
The complete quantum-classical hybrid self-attention computational flow integrates quantum feature transformation with classical attention weight computation:

\subsubsection{Query-Key-Value Triplet Generation}
\begin{equation}
    \mathbf{Q}^{(\ell)} = \mathcal{U}_{VQC}(\mathbf{X}^{(\ell)}; \Theta_{VQC}^{(\ell)}) \tag{21a}  
\end{equation}
\begin{equation}
    \\ \mathbf{K}^{(\ell)} = \mathbf{X}^{(\ell)} \mathbf{W}_K^{(\ell)} + \mathbf{b}_K^{(\ell)} \tag{21b}  
\end{equation}
\begin{equation}
    \mathbf{V}^{(\ell)} = \mathbf{X}^{(\ell)} \mathbf{W}_V^{(\ell)} + \mathbf{b}_V^{(\ell)}  \tag{21c}  
\end{equation}
where $\mathcal{U}_{\text{VQC}}(\cdot; \Theta_{\text{VQC}})$ represents the complete variational quantum circuit transformation pipeline (including dimensionality reduction, quantum encoding, Ansatz evolution, measurement, and upscaling), and $\Theta_{\text{VQC}}^{(\ell)}$ encompasses all trainable parameters related to quantum circuits for this layer. Key and Value maintain classical linear projections to balance computational efficiency with quantum enhancement benefits.

\subsubsection{Scaled Dot-Product Attention}
Computing attention score matrices based on quantum-generated Query and classical Key:
\begin{equation}
    \mathbf{A}^{(\ell)} = \text{softmax}\left(\frac{\mathbf{Q}^{(\ell)} (\mathbf{K}^{(\ell)})^T}{\sqrt{d_k}} + \mathbf{M}_{\text{mask}}\right) \tag{22}
\end{equation}
where $d_k = d_{\text{model}}/h$ is the dimension per attention head, $h$ is the number of heads, and $\mathbf{M}_{\text{mask}}$ is the causal mask matrix (for Decoder-Only architectures). Softmax normalization ensures attention weights $\mathbf{A}_{ij}$ satisfy probability distribution constraints $\sum_{j} A_{ij} = 1$.

The attention-weighted output is:
\begin{equation}
    \mathbf{O}_{\text{single}}^{(\ell)} = \mathbf{A}^{(\ell)} \mathbf{V}^{(\ell)} \in \mathbb{R}^{B \times L \times d_k} \tag{23}
\end{equation}

\subsubsection{Multi-Head Parallelization and Fusion}
Extending to $h$ independent quantum-classical hybrid attention heads, where each head $i \in \{1, \ldots, h\}$ maintains independent quantum circuit parameters $\Theta_{\text{VQC},i}^{(\ell)}$ and projection matrices $\{\mathbf{W}_{K,i}^{(\ell)}, \mathbf{W}_{V,i}^{(\ell)}\}$:
\begin{equation}
    \text{head}_i = \text{Attention}\left(\mathcal{U}_{\text{VQC},i}(\mathbf{X}^{(\ell)}), \mathbf{X}^{(\ell)}\mathbf{W}_{K,i}^{(\ell)}, \mathbf{X}^{(\ell)}\mathbf{W}_{V,i}^{(\ell)}\right) \tag{24}
\end{equation}
Multi-head outputs are aggregated through concatenation and linear transformation:
\begin{equation}
    \text{MultiHead}(\mathbf{X}^{(\ell)}) = \text{Concat}(\text{head}_1, \text{head}_2, \ldots, \text{head}_h) \mathbf{W}_O^{(\ell)} \tag{25}
\end{equation}
where $\mathbf{W}_O^{(\ell)} \in \mathbb{R}^{d_{\text{model}} \times d_{\text{model}}}$ is the output projection matrix. The multi-head mechanism enables the model to attend in parallel to different quantum feature subspaces of the input sequence, enhancing representation diversity and robustness.

\subsection{Deep Transformer Layer Residual Connections and Normalization}
The complete quantum-enhanced Transformer layer follows a Post-LayerNorm architecture, incorporating residual connections to mitigate gradient vanishing in deep networks:
\begin{equation}
    \tilde{\mathbf{X}}^{(\ell)} = \mathbf{X}^{(\ell)} + \text{Dropout}\left(\text{MultiHead}(\text{LayerNorm}(\mathbf{X}^{(\ell)}))\right) \tag{26a} 
\end{equation}
\begin{equation}
    \mathbf{X}^{(\ell+1)} = \tilde{\mathbf{X}}^{(\ell)} + \text{Dropout}\left(\text{FFN}(\text{LayerNorm}(\tilde{\mathbf{X}}^{(\ell)}))\right) \tag{26b}
\end{equation}
where the feedforward network FFN adopts a two-layer fully-connected structure:
\begin{equation}
    \text{FFN}(\mathbf{X}) = \text{GELU}(\mathbf{X}\mathbf{W}_1 + \mathbf{b}_1)\mathbf{W}_2 + \mathbf{b}_2 \tag{27}
\end{equation}
Layer normalization independently standardizes each sample, with computation formula:
\begin{equation}
    \text{LayerNorm}(\mathbf{x}) = \frac{\mathbf{x} - \mu}{\sqrt{\sigma^2 + \epsilon}} \odot \boldsymbol{\gamma} + \boldsymbol{\beta} \tag{28}
\end{equation}
where $\mu, \sigma^2$ are the mean and variance of that sample's features, $\boldsymbol{\gamma}, \boldsymbol{\beta}$ are learnable scaling and shift parameters, and $\epsilon$ is a numerical stability constant.

Stacking $N_{\text{layers}}$ such quantum-enhanced Transformer layers constitutes the complete quantum-classical hybrid LLM backbone network:
\begin{equation}
    \mathbf{H} = \mathcal{F}_{\text{Transformer}}(\mathbf{X}^{(0)}) = \left(\prod_{\ell=N_{\text{layers}}}^{1} \mathcal{F}_{\text{layer}}^{(\ell)}\right)(\mathbf{X}^{(0)}) \tag{29}
\end{equation}
where $\mathbf{X}^{(0)}$ is the input embedding layer output and $\mathbf{H}$ is the final hidden state representation.

\subsection{End-to-End Training and Quantum Gradient Computation}
The model employs an end-to-end supervised learning paradigm, jointly optimizing quantum circuit variational parameters $\Theta_{\text{VQC}}$ and classical neural network parameters $\Theta_{\text{classical}}$ through backpropagation algorithms. The training objective is to minimize the cross-entropy loss function for language modeling:
\begin{equation}
    \mathcal{L}(\Theta_{\text{VQC}}, \Theta_{\text{classical}}) = -\frac{1}{N_{\text{samples}}} \sum_{i=1}^{N_{\text{samples}}} \sum_{t=1}^{L_i} \sum_{k=1}^{V} y_{i,t,k} \log \hat{p}_{i,t,k} \tag{30}
\end{equation}
where $N_{\text{samples}}$ is the number of training samples, $L_i$ is the sequence length of the $i$-th sample, $V$ is the vocabulary size, $y_{i,t,k} \in \{0,1\}$ is the one-hot encoded ground truth label, and $\hat{p}_{i,t,k}$ is the model-predicted vocabulary probability distribution.

\subsubsection{Finite Difference Approximation for Quantum Gradients}
Due to the nonlinearity of quantum circuits and probabilistic nature of measurements, gradients of quantum parameters $\theta \in \Theta_{\text{VQC}}$ cannot be directly obtained through classical automatic differentiation. We employ the Central Finite Difference method for numerical gradient estimation:
\begin{equation}
    \frac{\partial \mathcal{L}}{\partial \theta_j} \approx \frac{\mathcal{L}(\theta_j + \delta) - \mathcal{L}(\theta_j - \delta)}{2\delta} \tag{31}
\end{equation}
where $\delta > 0$ is a small perturbation step size, practically chosen as $\delta \in [10^{-4}, 10^{-3}]$ to balance numerical precision and roundoff errors. Central differences possess $\mathcal{O}(\delta^2)$ truncation error accuracy compared to forward differences, but require two additional quantum circuit forward executions.

For quantum circuits containing $|\Theta_{\text{VQC}}|$ variational parameters, complete gradient computation requires $2|\Theta_{\text{VQC}}|$ circuit evaluations, with total computational complexity $\mathcal{O}(|\Theta_{\text{VQC}}| \cdot T_{\text{circuit}})$, where $T_{\text{circuit}}$ is the single circuit execution time. Despite introducing constant factors compared to the parameter-shift rule, the finite difference method's implementation simplicity and universality for arbitrary quantum gates provide advantages in practical deployment.

\subsubsection{Hybrid Optimization Strategy}
The Adam (Adaptive Moment Estimation) optimizer is employed for parameter updates, combining momentum methods with adaptive learning rate adjustments:
\begin{equation}
    \begin{aligned}
        \mathbf{m}_t &= \beta_1 \mathbf{m}_{t-1} + (1-\beta_1) \mathbf{g}_t \\
        \mathbf{v}_t &= \beta_2 \mathbf{v}_{t-1} + (1-\beta_2) \mathbf{g}_t^2 \\
        \hat{\mathbf{m}}_t &= \mathbf{m}_t / (1-\beta_1^t) \\
        \hat{\mathbf{v}}_t &= \mathbf{v}_t / (1-\beta_2^t) \\
        \boldsymbol{\theta}_t &= \boldsymbol{\theta}_{t-1} - \eta_t \frac{\hat{\mathbf{m}}_t}{\sqrt{\hat{\mathbf{v}}_t} + \epsilon}
    \end{aligned} \tag{32}
\end{equation}
where $\mathbf{g}_t$ is the gradient at step $t$, $\mathbf{m}_t, \mathbf{v}_t$ are first and second moment estimates of gradients respectively, $\beta_1=0.9, \beta_2=0.999$ are exponential decay rates, and $\epsilon=10^{-8}$ is a numerical stability term. Bias correction terms $\hat{\mathbf{m}}_t, \hat{\mathbf{v}}_t$ compensate for bias in moment estimates during initial stages.

\subsubsection{Cosine Annealing Learning Rate Schedule}
The learning rate $\eta_t$ adopts a Cosine Annealing with Warm Restarts strategy:
\begin{equation}
    \eta_t = \eta_{\min} + \frac{1}{2}(\eta_{\max} - \eta_{\min})\left(1 + \cos\left(\frac{T_{\text{cur}}}{T_{\max}} \pi\right)\right) \tag{33}
\end{equation}
where $\eta_{\max}$ is the maximum learning rate (initial learning rate), $\eta_{\min}$ is the minimum learning rate, $T_{\text{cur}}$ is the current training step within the period, and $T_{\max}$ is the total steps in a single annealing cycle. This strategy smoothly decays the learning rate from $\eta_{\max}$ to $\eta_{\min}$ during training, then restarts to $\eta_{\max}$ periodically, facilitating escape from local optima and exploration of broader parameter spaces—particularly effective for the non-convex optimization landscape of quantum-classical hybrid systems.

Linear warmup is employed in early training, linearly increasing the learning rate from 0 to $\eta_{\max}$ over the first $T_{\text{warmup}}$ steps:
\begin{equation}
    \eta_t = \eta_{\max} \cdot \frac{t}{T_{\text{warmup}}}, \quad t \leq T_{\text{warmup}} \tag{34}
\end{equation}
The warmup phase mitigates training instability caused by improper quantum parameter initialization, establishing a solid foundation for subsequent optimization.

\section{Experiments}

\subsection{Experimental Setup}
To rigorously evaluate the performance, efficiency, and stability of HyQuT, we conducted a series of experiments across two distinct model scales. We instantiated two primary hybrid model configurations, each targeting a different strategic location within the Transformer architecture, and compared them against their corresponding pure classical baselines. The first configuration, HyQuT-8M, was developed on an 8-million-parameter scale model where the VQC module replaces the gate projection matrix within the Feed-Forward Network. The second, HyQuT-150M, was implemented on a larger 150-million-parameter model, with the VQC module replacing the Query ($W_q$) projection matrix in the self-attention block. These two configurations allowed us to assess our framework's viability for both rapid prototyping and in more realistic, large-scale settings. The key hyperparameters for their respective classical baselines, Classic-8M and Classic-150M, are detailed in Table \ref{tab:Hyperparameters}.

The models were pre-trained on a 1.6 GB Chinese text corpus, which was curated from the Jiangshu Large Model Dataset \cite{deepctrl_sft_dataset} and filtered for sequences shorter than 512 characters. The pre-training was conducted on a single server equipped with eight H20 GPUs, using a batch size of 32 for a total of 5 epochs (27,585 steps). To ensure stable convergence, we employed a learning rate schedule featuring an initial warm-up phase, a stable plateau, and a final cosine decay.

To provide a comprehensive assessment, our evaluation was structured around three key criteria. First, we measured Resource Efficiency by quantifying the total number of trainable parameters and the theoretical computational cost in Floating-Point Operations (FLOPs). Second, we analyzed the Training Dynamics by examining the pre-training loss curve of each model to evaluate its convergence behavior and stability. Finally, to determine practical performance, we assessed the Generative Quality of the models through a qualitative analysis of their text generation capabilities, focusing on the coherence, context-awareness, and overall quality of the responses.

\begin{table}
 \caption{Hyperparameters of Classical Baseline Models}
  \centering

  \begin{tabular}{lrr}
    \toprule
    \textbf{Hyperparameter}     & \textbf{Classic-8M}     & \textbf{Classic-150M} \\
    \midrule
    hidden\_size & 512  & 1024     \\
    num\_hidden\_layers     & 2 & 16      \\
    num\_attention\_heads     & 8       & 8  \\
    num\_key\_value\_heads & 2  & 2     \\
    intermediate\_size     & 1024 & 2048      \\
    max\_position\_embeddings     & 4096       & 32768  \\
    seq\_len & 512  & 512     \\
    \bottomrule
  \end{tabular}
  \label{tab:Hyperparameters}
\end{table}

\subsection{Ablation Study on Replacement Strategy}
To identify the most effective and stable integration points for the VQC module, we conducted a comprehensive ablation study on the Classic-8M model. The primary goal of this investigation was to understand the sensitivity of the Transformer architecture to the replacement of its various linear layers and to determine the threshold beyond which training stability is compromised. This study allowed us to systematically map out the trade-offs between parameter reduction and model convergence, providing an empirical basis for our main experimental design.

The results of this study are summarized in Table \ref{tab:ablation_study}. We systematically replaced individual and combined linear layers within both the self-attention and feed-forward network blocks and monitored the model's ability to converge. The parameter count and the estimated Floating-Point Operations (FLOPs) were also calculated for each configuration to quantify the resource savings. To provide a reasonable estimate for the HyQuT-8M (Gate) configuration, its FLOPs were approximated by substituting the VQC with a classical dense layer of equivalent input-output dimensions.

\begin{table}
 \caption{Ablation Study on VQC Replacement Strategies in HyQuT-8M}
  \centering
  \resizebox{\linewidth}{!}{
  \begin{tabular}{lrrrc}
    \toprule
    \textbf{Replacement Target} & \textbf{Parameters (M)} & \textbf{FLOPs (B)} & \textbf{Classical Computation (\%)} & \textbf{Converged} \\
    \midrule
    None (Classical Baseline) & 7.748 & 760.3 & 100.00 & $\surd$ \\
    Attention: $W_q$             & 7.246 & 710.8 & 93.49  & $\surd$ \\
    Attention: $W_q$, $W_k$, $W_v$     & 7.032 & 689.1 & 90.64  & $\surd$ \\
    Attention: $W_q$, $W_k$, $W_v$, $W_o$ & 6.525 & 639.7 & 84.14  & -       \\
    \midrule 
    FFN: $W_{gate}$                & 6.722 & 657.5 & 86.48  & $\surd$ \\
    FFN: $W_{gate}$, $W_{up}$, $W_{down}$    & 4.690 & 459.6 & 60.45  & -       \\
    \midrule 
    All Linear Layers         & 3.466 & 339.0 & 44.59  & -       \\
    \bottomrule
  \end{tabular}
  }
  \label{tab:ablation_study}
\end{table}

As the results in Table \ref{tab:ablation_study} clearly indicate, a distinct pattern emerged regarding training stability. The model successfully converged when individual, functionally critical matrices such as the Query ($W_q$) or the FFN's Gate ($W_{gate}$) were replaced. Even a combined replacement of the Query, Key, and Value matrices in the attention block maintained convergence. However, more aggressive strategies that targeted the entire set of linear layers within a functional block—such as all four attention matrices or all three FFN matrices—consistently led to training instability and a failure to converge. This empirical evidence suggests that while the Transformer architecture is robust enough to accommodate a targeted, 'surgical' integration of quantum circuits, its core computational mechanisms are sensitive to large-scale, simultaneous substitutions. These findings strongly informed our decision to focus on the two most promising and stable configurations—replacing the FFN Gate for the HyQuT-8M model and the Attention Query for the HyQuT-150M model—for our main, in-depth analysis.

\subsection{Core Results and Analysis }
Following the experimental plan, we conducted a detailed analysis of the hybrid models' performance against their classical baselines. The core results confirm that our hybrid quantum-classical framework can achieve significant resource savings without compromising training stability or the ultimate quality of language generation.

A primary motivation for this work is to reduce the classical computational resources required by LLMs. We quantified this by comparing the parameter counts and theoretical FLOPs of our hybrid models against their classical counterparts. As detailed in Table \ref{tab:resource_efficiency}, both hybrid configurations demonstrated notable efficiency gains. The HyQuT-8M model, for instance, lowered the classical computational workload of the 8M-scale baseline by over 13.5\%, with a corresponding decrease in classical FLOPs. For a complete, layer-by-layer breakdown of the parameter distribution along with the memory usage within the HyQuT-8M model, providing full architectural transparency, please refer to Appendix A.

\begin{table}
 \caption{Resource Efficiency Comparison for Hybrid and Classical Models}
  \centering
  \resizebox{\linewidth}{!}{
  \begin{tabular}{lrrr}
    \toprule
    \textbf{Model Configuration} & \textbf{Parameters (M)} & \textbf{Parameter Reduction (\%)} & \textbf{FLOPs (T)} \\
    \midrule
    Classic-150M & 149  & -     & 14.66  \\
    HyQuT-150M   & 133  & 10.7  & 13.02  \\
    \addlinespace 
    Classic-8M   & 7.75 & -     & 0.76   \\
    HyQuT-8M     & 6.72 & 13.3  & 0.66  \\
    \bottomrule
  \end{tabular}
  }
  \label{tab:resource_efficiency}
\end{table}

\begin{figure}
  \centering
  \includegraphics[width=0.6\textwidth]{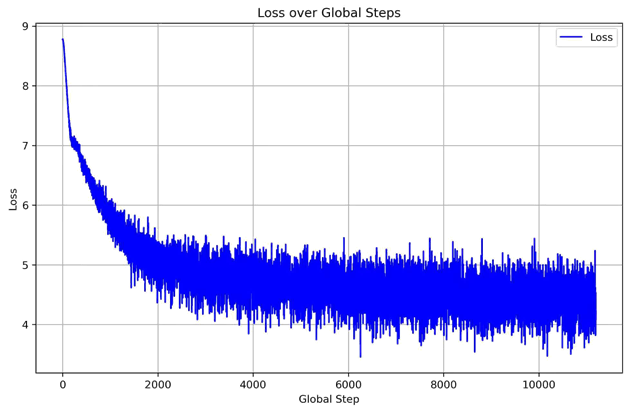}
  \caption{Loss Curve of HyQuT-8M}
  \label{fig:8m-loss}
\end{figure}

\begin{figure}
  \centering
  \includegraphics[width=0.6\textwidth]{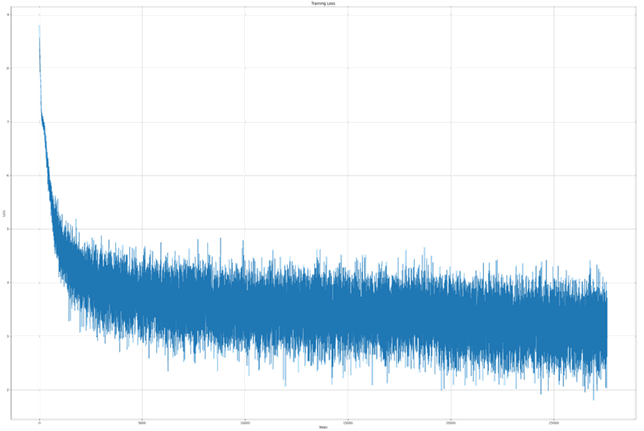}
  \caption{Loss Curve of HyQuT-150M}
  \label{fig:150m-loss}
\end{figure}

A critical test for any hybrid architecture is its ability to integrate into the training process and achieve stable convergence. We monitored the pre-training loss curves for both our hybrid models to assess this stability, with the results presented in Figure \ref{fig:8m-loss} and Figure \ref{fig:150m-loss}. Both models demonstrated smooth and successful learning trajectories, confirming the viability of our approach.

At the 8M scale, the HyQuT-8M model exhibits a consistent decrease in loss, starting from an initial value of approximately 8.0 and steadily converging to a final value near 4.0. This positive result was successfully replicated at a much larger scale. The HyQuT-150M model also shows excellent convergence stability. As would be expected for a model with significantly greater capacity, it achieves an even lower final loss, converging to a value of approximately 3.0.

The successful and stable convergence of both models provides strong evidence that our hybrid module integrates seamlessly into the standard backpropagation-based optimization process. The fact that the framework facilitates effective learning without disruption at both an 8M and a 150M scale is a crucial finding. It demonstrates that our approach is robust and scalable, suggesting its viability for developing even larger hybrid models in the future.

While quantitative metrics for efficiency and stability are essential, the ultimate measure of a language model's success is the quality of the text it generates. To evaluate the practical capabilities of our hybrid architecture, we performed a qualitative analysis of the text generated by the pre-trained HyQuT-150M model. The model was prompted with a variety of inputs designed to test its ability to handle different conversational and explanatory tasks.

\begin{figure}
  \centering
  \includegraphics[width=1.0\textwidth]{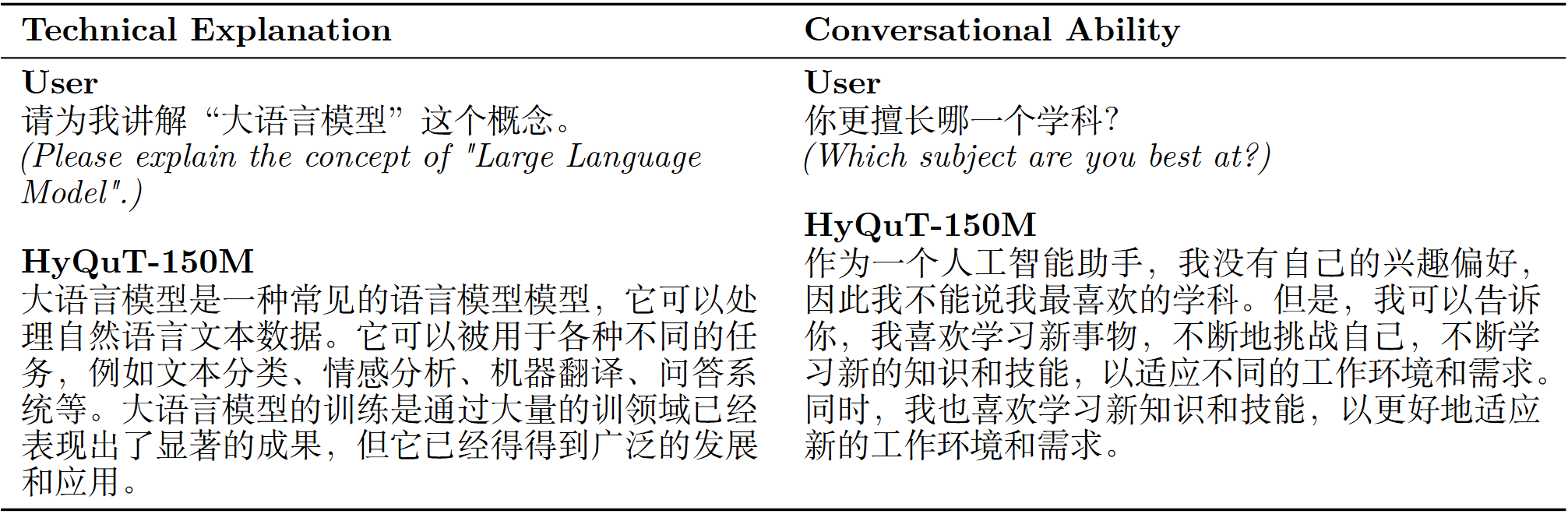}
  \caption{Qualitative Examples of Text Generated by the HyQuT-150M Model}
  \label{tab:qualitative_examples}
\end{figure}

Representative examples of the generated text are presented in Figure \ref{tab:qualitative_examples}. The outputs demonstrate that the HyQuT-150M model successfully learns foundational language capabilities. It correctly interprets the user's intent, maintains contextual relevance, and generates multi-sentence responses that are coherent and on-topic. 

\section{Conclusion}
In this work, we introduced HyQuT, the first hybrid quantum-classical large language model demonstrated to be capable of coherent, generative NLP. By successfully integrating a variational quantum circuit into Transformer architectures—implemented as HyQuT-8M and HyQuT-150M—our research bridges the gap between the theoretical promise of quantum computing and the practical demands of large-scale generative AI. This study moves beyond the established domain of quantum machine learning for classification and regression tasks, venturing into the more complex and challenging realm of language generation.

Our experimental results provide a strong proof-of-concept for the viability of this hybrid approach. We have shown that a remarkably compact quantum component—requiring only 10 qubits and approximately 80 quantum gates—can effectively replace over 10\% of the classical parameters in a 150M-scale model. Crucially, this significant reduction in classical resources was achieved without compromising the model's training stability, as evidenced by its consistent loss convergence. The preservation of the model's core ability to generate contextually relevant text further underscores the feasibility of our framework.

While our findings are promising, we acknowledge that this work represents an initial but crucial step. A natural and important direction is to scale this methodology to larger and more capable architectures. Furthermore, investigating the integration of quantum modules into different model families could reveal new opportunities for efficiency gains. Ultimately, this research paves the way for a new paradigm of quantum-enhanced NLP, offering a potential path to mitigate the escalating computational demands of classical LLMs.

\section*{Acknowledgments}
Part of this work is supposed by the Yangtze Delta Industrial Innovation Center of Quantum Science and Technology, Suzhou.

\bibliographystyle{unsrt}  
\bibliography{references}  

\newpage
\appendix

\section{Detailed Parameter Breakdown of the HyQuT-8M Model}
\label{app:param_breakdown}

\begin{table}[htpb]
 \centering
 \begin{threeparttable}
  \caption{Layer-by-layer parameter distribution of the HyQuT-8M model.}
  \label{tab:param_breakdown}
  \begin{tabular}{l l r r}
    \toprule
    \textbf{Model Layer Name} & \textbf{Dimensions} & \textbf{Parameter Count} & \textbf{Memory (MB)} \\
    \midrule
    model.embed\_tokens.embedding\_table & (6401, 512) & 3,277,312 & 12.50 \\
    \addlinespace
    \multicolumn{4}{c}{\textit{Transformer Layer Block (Structure repeats for 2 layers)}} \\
    \addlinespace
    layers.N.input\_layernorm.weight & (512,) & 512 & <0.01 \\
    layers.N.self\_attn.q\_proj.weight & (512, 512) & 262,144 & 1.00 \\
    layers.N.self\_attn.k\_proj.weight & (128, 512) & 65,536 & 0.25 \\
    layers.N.self\_attn.v\_proj.weight & (128, 512) & 65,536 & 0.25 \\
    layers.N.self\_attn.o\_proj.weight & (512, 512) & 262,144 & 1.00 \\
    layers.N.post\_attention\_layernorm.weight & (512,) & 512 & <0.01 \\
    \addlinespace
    layers.N.mlp.gate\_proj.* (Hybrid Module)  & & & \\
    ...reduce\_proj.weight & (20, 512) & 10,240 & 0.04 \\
    ...reduce\_proj.bias & (20,) & 20 & <0.01 \\
    ...mq\_layers.0.weight (Quantum Params) & (40,) & 40 & <0.01 \\
    ...dense\_expand.weight & (1024, 1) & 1,024 & <0.01 \\
    ...dense\_expand.bias & (1024,) & 1,024 & <0.01 \\
    \addlinespace
    layers.N.mlp.down\_proj.weight & (512, 1024) & 524,288 & 2.00 \\
    layers.N.mlp.down\_proj.bias & (512,) & 512 & <0.01 \\
    layers.N.mlp.up\_proj.weight & (1024, 512) & 524,288 & 2.00 \\
    layers.N.mlp.up\_proj.bias & (1024,) & 1,024 & <0.01 \\
    \addlinespace
    \multicolumn{4}{c}{\textit{End of Transformer Layer Block}} \\
    \addlinespace
    model.norm.weight & (512,) & 512 & <0.01 \\
    lm\_head.bias & (6401,) & 6,401 & 0.02 \\
    \midrule
    \textbf{Total} & - & \textbf{6,721,913} & \textbf{25.64 MB} \\
    \bottomrule
  \end{tabular}
  \begin{tablenotes}[para,flushleft]
    \small 
    \item[*] Note: The \texttt{mlp.gate\_proj} block details the components of our Hybrid Quantum Module. \texttt{mq\_layers.0.weight} represents the trainable parameters $\theta$ of the VQC.
  \end{tablenotes}
 \end{threeparttable}
\end{table}

\end{document}